# Artificial Intelligence for the Characterization of Particles and Fibers by Optical Microscopy

Simiao Sun[1], Kenneth Ng[1], Lynn Lee[2], Astrid Harth[3], Asami Odate[4], Aggelos Katsaggelos[5], Manuel Ballester Matito[5], Nicholas Eastaugh[6], Marc Walton[2*]

*Corresponding Author: mwalton@hku.hk

## Abstract

Optical microscopy of particle and fiber dispersions involves interpreting subtle visual cues influenced by specimen morphology, chemical composition, magnification, and illumination conditions. We introduce an artificial intelligence (AI) distillation framework that extracts semantically rich image embeddings from microscopy images using semantic anchors. A multimodal teacher combines each image's visual embedding with three text embeddings representing illumination modality, magnification, and specimen identity and morphology. Generated by LongCLIP's extended-context text encoder, this yields a 2304-dimensional block-structured teacher vector whose component blocks remain physically interpretable throughout training and inference. A student vision transformer (ViT) with a multi-layer perceptron (MLP) decoder is trained to reconstruct this teacher vector from the image alone, minimizing a mean absolute error (L1) loss that enforces coordinate-level fidelity to the teacher's block structure. A cross-entropy term over pseudo-classes derived from HDBSCAN clustering of the teacher embedding space acts as a collapse-prevention regularizer, enforcing inter-cluster separation without requiring contrastive negative mining. At inference, the student operates on image input alone, producing compact embeddings that recover the full semantic content of the teacher vector. The framework achieves approximately 80% pseudo-class validation accuracy and 75% Recall@1 on fine-grained specimen description labels under leave-one-out nearest-neighbor retrieval. These results demonstrate that semantic anchoring enables a vision-only student to acquire richer and more interpretable representations than image-only training, with direct applicability to retrieval, classification, and exploratory analysis of heterogeneous particle and fiber dispersions.

## 1.0 Introduction

Since the pioneering studies of Laurie, [1] Gettens and Stout, [2] and McCrone et al., [3] microscopy of particles, and particularly under polarized light (as in polarized light microscopy or PLM), has played a central role in the analysis of cultural heritage materials, and has persisted as a technique for characterization even as more advanced analytical techniques have become available. [4,5] Compared with other instrumental methods used in cultural heritage, such as scanning electron microscopy (SEM), Fourier transform infrared spectroscopy (FTIR), X-ray fluorescence (XRF), and Raman spectroscopy, optical microscopy still offers highly specific and sensitive material identification. Its key advantage is accessibility and affordability, making it a practical choice for both conservators and scientists. Often underestimated, optical microscopy is a powerful ultra-microanalysis tool and can exceed the detection capabilities of even the most advanced analytical methods, particularly in its ability to image and characterize particles in the

1) Department of Chemistry, University of Hong Kong, Hong Kong SAR
2) Museum Studies Program, University of Hong Kong, Hong Kong SAR
3) Department of History, City University, Hong Kong, Hong Kong SAR
4) M+ Museum of Visual Culture, Hong Kong, Hong Kong SAR
5) Department of Electrical and Computer Engineering, Northwestern University, USA
6) Visarik, LLC, London, United Kingdom

micron-to-submicron range.[6] As written by McCrone in the *Journal of the American Institute of Conservation,*[7]

> *After microscopic enlargement, most tiny pigments, fibers, etc., are usually identified by a process all of us use thousands of times a day to recognize people, buildings, trees, books, cars, etc. Many microscopic substances enlarged 10-1000x are then recognized just as definitely, and just as rapidly, as we recognize macroscopic objects without magnification.*

This quote concisely explains why optical microscopy has held a singular status in art conservation as an approachable method of analysis, conceptually similar to other common visual assessment methods, like close looking,[8] but at the microscale. The practice of microscopy in conservation has resulted in the development of collections of particle and fiber images for comparative purposes. This study primarily focuses on these images compiled from disparate reference databases. We are interested in how these legacy images can be reused in today's world that has become increasingly reliant on artificial intelligence (AI) for recognizing "*people, buildings, trees, books, cars, etc.*" [9]

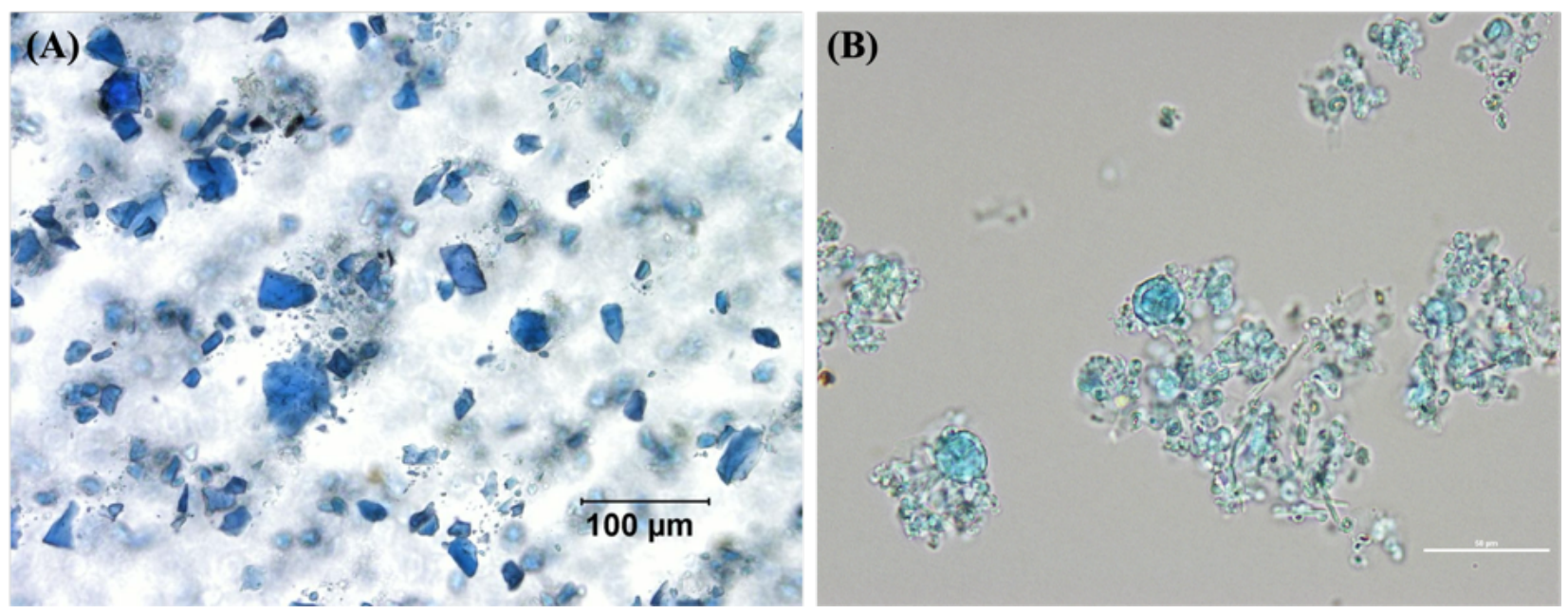

***Figure 1**:* Two morphological forms of azurite ($Cu_3(CO_3)_2(OH)_2$): (A) natural azurite (note scale bar is 100 μm), (B) synthetic azurite (note scale bar is 50 μm), which can also be known by the common names of Blue Bice, Blue Verditer, Bremen Blue, etc. Also note the contrast difference with the background despite both being plane-polarized images. Also, the particle sizes are on two different scales, which is a limitation in this combined dataset of legacy data.

Images of microparticles from the limited published datasets [10-12] include scanned archival photomicrographs and natively digital files of varying sizes and qualities. A brief survey of these images shows that the appearance of chemically identical samples can vary widely due to factors such as morphology, particle size, or illumination conditions. Conversely, visually similar images may represent entirely different materials, often due to inconsistencies in image acquisition, particularly given the diverse range of microscopes and, moreover, the analysts who produce the photomicrographs. This is not a unique problem to this field, and in computer vision applications, there are similar problems with recognizing objects due to pose, lighting conditions, zoom factors, etc., but the problem is not as pronounced as that seen in Figure 1.

Also, the metadata associated with these microparticle images is a mix of technical, chemical, and historical terminology, with little uniformity or standardization. Known as "weak" or "pseudo" labels in deep learning, these imprecise or inconsistent metadata hinder the development of reliable AI models.[13,14] In cultural heritage, a clear example of such a weak label are the numerous names of the mineral azurite, which is also known as copper carbonate, blue verditer, blue bice, Bremen's blue, *etc.*, which all refer to the same molecular substance but are named differently—either as mineral names, chemical descriptions, or common names for the synthetic versions of blue pigment, but often appear as separate entries in respective databases.[10] This variability means that even correctly labeled samples— in terms of historic contexts or time periods of manufacture of the material— might not align with unified scientific concepts, such as the blue colored basic copper carbonate's chemical formula, $Cu_3(CO_3)_2(OH)_2$, its characteristically strong birefringence (~0.110), or that the synthetic version of the pigment has a morphology that appears very different from the natural substance (Figure 1). Indeed, weakly labeled or entirely unlabeled data is not unique to cultural heritage but is a universal challenge and an active research frontier, especially in self-supervised vision transformers (*e.g.*, BYOL, JEPA, DINO), which aim to reduce dependence on manual annotation while advancing toward more general visual intelligence.[15-17]

Efforts to systematize the capture and search of particle and fiber images with defined features and definitions are not new. In the 1970s, McCrone's *The Particle Atlas* aimed to standardize image capture with "partially crossed" polarization filters with their PLM data designed to maximize visual information and included detailed descriptions, though its utility today is limited by inconsistencies in the tonalities and fading of offset color printing and the lack of structured data within the volume.[12] Moreover, the McCrone Institute's online database, a Web 1.0 extension of the Particle Atlas, was removed from its site in 2017, with the expectation that the company would relaunch the site in a refashioned form soon (but as of publication, this has not yet appeared). The *Pigment Compendium,* when it was published in 2008, expanded the amount of culturally relevant pigment particle data by using CD-ROM technology, which required state-of-the-art digital capture of reference standards and included searchable, structured information for each entry.[10] While revolutionary at the time, the *Pigment Compendium* was mainly designed for manual interpretation and comparison rather than computational analysis. The same is true for the Museum of Fine Arts, Boston's CAMEO database,[11] which was designed as a Web 2.0 Wiki page first created in the early 2000s, but now lacks built-in capabilities for semantic cross-referencing inherent to the latest AI advances.

Our motivation is to build on these former contributions to automate microscopic image retrieval with the state-of-the-art deep image networks, like vision transformers (ViTs). This problem, where in a query image is introduced to the system to find the most similar images in a database remains challenging in microscope contexts. ViTs trained on ImageNet can extract natural features and color contrasts but are not natively able to distinguish subtle features common in microscopy, such as background color shifts or polarization variations, which are attributed to the specimen's chemical or mineralogical properties.[18,19] Also, magnification and size inconsistencies can distort the model's ability to distinguish features at consistent length scales. Consequently, computer vision models are often constrained by a fundamental disconnect between the data on which they are trained and the specialized data they are meant to interpret, known as a domain mismatch or shift.[20]

The domain shift in this study is that the natural images used to train foundation models, lack the same statistical contrast and feature relationships found in microscopic images.[21] Due to these issues, machine learning approaches have yet to match the nuanced skills of an experienced human optical microscopist. Moreover, many of these domain-specific legacy microscopic datasets, especially in cultural heritage, lack the structured labels and metadata required for supervised learning, and, most importantly, their labeling cannot easily be outsourced to non-experts through labeling services, like Amazon Mechanical Turk (MTurk), due to their highly specialized scientific nature.[22] This means that if we are to advance microscopy with artificial intelligence, self-supervision or semi-self-supervision approaches are necessary.

Here, we introduce a framework to maximize the utility of legacy cultural-heritage data, often sourced from multiple institutions and collected over the past 60 years. Specifically, we assemble a weakly labeled corpus of microparticle images and associated metadata, pass them through a pretrained vision transformer and language encoder, and extract latent representations for both modalities. As the multimodal foundation, we use OpenAI's Contrastive Language-Image Pre-training model (CLIP),[23] which was jointly trained with vision and language encoders so that matched image-text pairs map to nearby points in a shared embedding space. Our key insight is that, because these embeddings are geometrically pre-aligned,[24] CLIP image and text representations can be concatenated to produce richer features that capture both visual appearance and semantic descriptions.[25] We hypothesize that fusing visual and textual features, even when metadata are inconsistent, can guide the training of a new vision model whose outputs encode material concepts as well as image contrast, rather than either signal alone.[26-29] Using semantic anchoring from the text encoder, meaningful cluster structure can still emerge despite annotation variability.[30-32] Leveraging this structure, visualized with Uniform Manifold Approximation and Projection (UMAP) [33,34] and clustered with Scikit-learn's Hierarchical Density-Based Spatial Clustering of Applications with Noise (HDBSCAN), we develop a semi-self-supervised student-teacher architecture in which a student network learns to reproduce the teacher's concatenated image-text embedding. [35-37] This data-driven approach yields a domain-adapted ViT for particle-image retrieval and provides a proof of concept for utilizing heterogeneous legacy data for materials characterization.

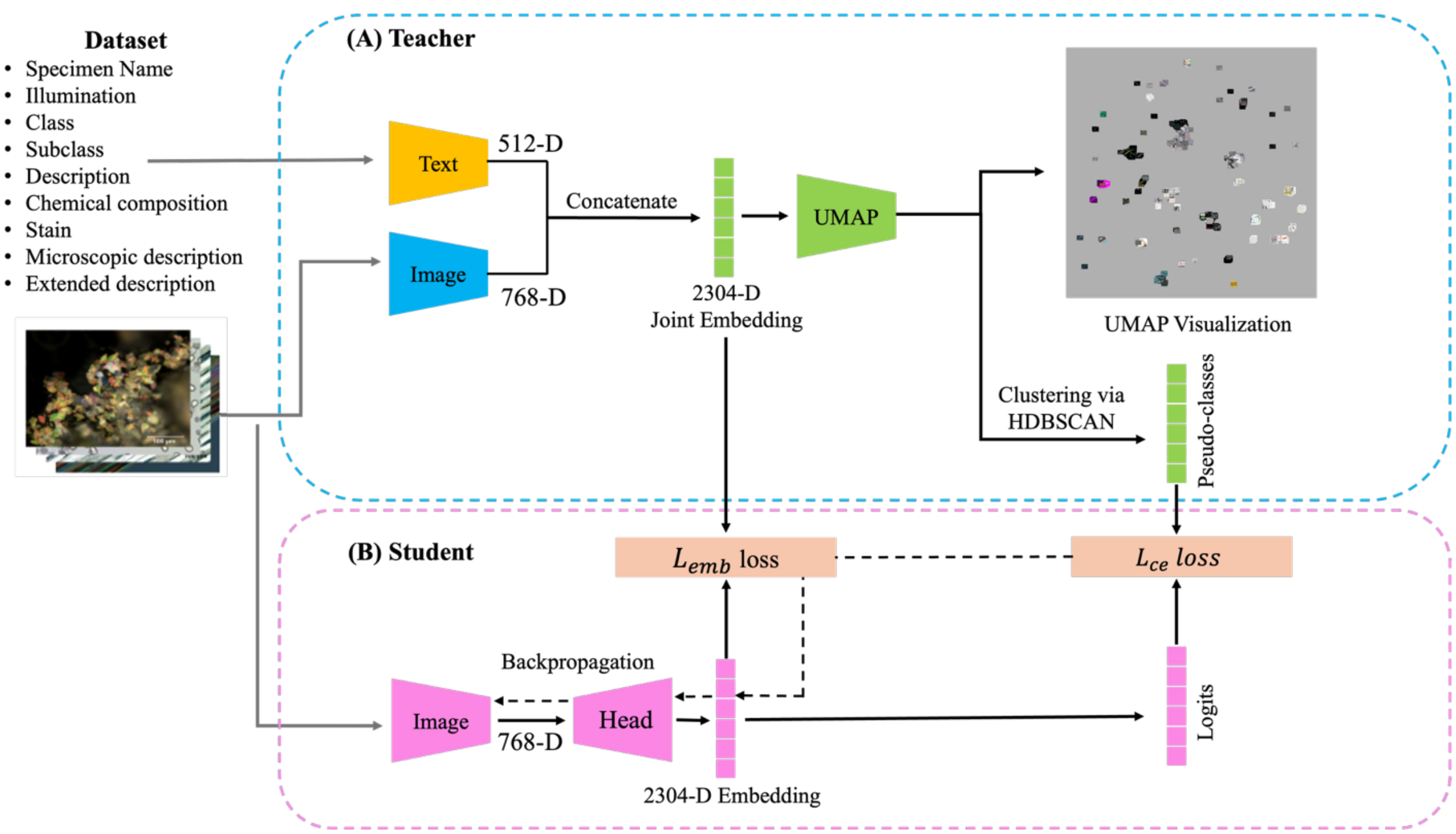

***Figure 2:*** Distillation framework. (A) Teacher branch: multimodal inputs (image + text) are encoded to form the fused teacher manifold. (B) Student branch: image-only inputs are trained to reproduce the teacher manifold.

# 2.0 Methods

## 2.1 *Description of Knowledge Distillation Pipeline*

The teacher-student distillation pipeline in this work (Figure 2) is inspired by Barlow Twins [38], which trains on two distorted views of the same image, each passed through a twin ViT encoder. In Barlow Twins, invariance (same teacher/student representations) is encouraged by pushing the diagonal entries of the cross-correlation matrix constructed from the output of these twin ViT's toward 1, while redundancy (diverse and independent representations) is reduced by pushing the off-diagonal entries toward 0, so the matrix approaches the identity. In the case of a collapsed constant representation, all entries in the correlation matrix would degenerate to 1, which strongly violates the off-diagonal penalty and prevents this scenario from occurring. This simple regularization mechanism was a key step in stabilizing self-supervised visual representation learning and influenced subsequent families of methods, including DINO- and JEPA-style approaches.

In our distillation adaptation, we are motivated by the same principle but achieve invariance by minimizing a mean absolute error (MAE) or $l_1$ reconstruction loss between student and teacher embeddings. This is a stricter constraint than Barlow's diagonal objective because it enforces pointwise agreement in absolute coordinates for each sample ($z_i \approx t_i$ in every dimension, where $z_i$ and $t_i$ denote student and teacher latent representations), whereas Barlow's diagonal term enforces only batch-level self-correlation invariance ($C_{jj} \rightarrow 1$) without uniquely anchoring each sample to a fixed target. The complementary redundancy-reduction role in our formulation is provided by a weak-label cross-entropy term that promotes class-discriminative partitioning of the embedding space. The total objective to be minimized is

$$L_{total} = L_{l1} + \beta L_{ce} \quad (1)$$

where $L_{l1}$ is the embedding reconstruction loss, $L_{ce}$ is weak-label cross-entropy, and $\beta > 0$ is a single relative weight. One advantage of our approach is that by interpreting the cross-entropy (CE) term as a surrogate for how well the student's latent representations encode the teacher's multimodal structure, [39] our formulation provides a robust alternative to Barlow-style redundancy control. In optimizing the training, we found that setting $\beta$ to a low value (0.11) best enforces the desired geometric alignment between teacher and student embeddings without approaching collapse into a redundant solution.

***Teacher Model.*** The teacher model in this framework (Figure 2a) is designed to enhance the cluster structure in the training set by anchoring image-level contrast cues to metadata-specific semantics. We use LongCLIP (ViT-B/16), implemented in PyTorch, to extract image and text embeddings from a backbone pretrained on large-scale natural-image corpora, and to fuse these modalities into a structured teacher manifold that serves as the distillation target for the student [30]. An advantage of LongCLIP is its 248-token text context window, compared with CLIP's 77 tokens. This additional capacity allows us to encode the extended microscopy descriptions used by specialists and incorporate them as semantic priors for feature identification. Another advantage of this ViT-B/16 variant is its relatively small size and 86M parameters, which means it can be trained without strict memory constraints given our access to standard GPU hardware.

As is standard for a ViT, in the vision encoder streams, the image is segmented into a sequence of fixed-size 16 x 16-pixel patches, which are then flattened, defined as $V = \{v_{cls}, v_1, \cdots, v_n\}$, where $v_{cls}$ is a learnable token used for global image representation, and $v_i$ denotes the flattened spatial patches capturing contrast features of interest. Second, in the text encoder stream, the metadata is tokenized into a sequence $T = \{t_1, \cdots, t_L\}$, where the element $t_i$ represents a semantic token (word or subword) and the sequence length of this token is capped at 248. Both streams transform their input tokens into embeddings, respectively 768 (extracted from the ln_post CLS layer) or 512-dimensional numerical vectors ($E_{image} \in \mathbb{R}^{768}, E_{text} \in \mathbb{R}^{512}$) as the model output.

We observed that embeddings were better separated when text fields were structured and encoded independently in the text stream, then averaged and concatenated with the corresponding image embedding for each sample. As will be detailed further below (in Implementation Details), under this formulation, teacher inference yields a 2304-dimensional fused embedding per sample. Across all 5410 images in the database, these embeddings are projected onto a two-dimensional UMAP manifold that preserves local neighborhoods while retaining the global structure of the dataset for visualization. The resulting HDBSCAN groupings of the latent embeddings are then used to derive pseudo-labels for the CE term in Eq. (1).

Unlike K-means, which forces every sample into a cluster, HDBSCAN infers clusters from local density and explicitly marks low-density samples as noise, allowing it to clearly identify materially homogenous groups. Using this clustering approach enables the fusion pipeline to produce compact, high-density groupings of distinct material types across illumination conditions, while appropriately isolating samples with ambiguous contrast signatures or weak metadata as noise. These noise points were excluded from student training. With this approach, the resulting cluster assignments can be interpreted as a discrete approximation to the joint image-text density induced by CLIP embeddings. Consequently, the pseudo-labels from such HDBSCAN clusters are data-driven rather than ontology-driven—an important property for heterogeneous, inconsistently annotated collections, where imposed taxonomies tend to either over-fragment visually similar materials or merge perceptually distinct ones.

***Student Model.*** The student model (Figure 2B) reuses the teacher ViT visual encoder as a feature extractor. In the initial training stage, the ViT backbone is frozen, and only the downstream multilayer perceptron (MLP) head and classifier are updated. Fused teacher embeddings provide a fixed reconstruction target for the $L_{l1}$ term of Eq. 1. To match the teacher embedding dimensionality, a decoder-style MLP progressively expands the student representation (768 → 1024 → 1536 → 2304), with the final layer aligned to the teacher's 2304-dimensional latent space. The first three blocks follow a linear → layer normalization → GELU → dropout (0.2) pattern, followed by a final linear classifier mapping latent embeddings to material-class logits.

Instead of employing a Barlow Twins-style covariance penalty, we mitigate redundancy using a similar constraint through CE on the weak material labels derived from HDBSCAN. The image-text embeddings within each cluster are assigned the cluster's label. Here, the CE term acts as a regularizer for the reconstruction goal, similarly to Barlow's off-diagonal penalty function. In essence, like Barlow Twins, this objective discourages trivial solutions with constant features without enforcing a specific non-trivial geometric configuration.

The CE gradient also guides the latent space to maintain the atomized structure already existing in the combined vision-language manifold. Per-sample CE on these pseudo-labels functions as a discriminative anchor, helping to organize the embedding space and enforce decision boundaries that align with the different material identities in our dataset. This approach sharpens clusters that might otherwise remain overly smooth when trained only on reconstruction, and directs gradients toward rare or subtle classes, where matching the teacher's output can be ambiguous or noisy.

Finally, the CE provides a monitor of the student's manifold quality during training. Accuracy on a held-out validation split reflects both the non-degeneracy of the learned embedding and student-teacher alignment. Relative to Barlow Twins, we therefore trade the broader information-theoretic generality of cross-correlation regularization for an objective that is tractable at our batch size and more directly aligned with our domain-specific goal of generating an image embedding with semantic richness.

### *2.2 Implementation Details*

***Construction of the Database.*** To address the challenges posed by limited training data, [40,41] we sought to construct the most comprehensive dataset possible by combining published and publicly available optical microscopy datasets either directly from cultural heritage or allied fields (see supplementary). This unified data was aggregated from the Museum of Fine Arts Boston's CAMEO database together with its Fiber Reference Image Library (FRIL) [11], The Pigment Compendium, [10] The Atlas of Microscopic Particles, [12] the University of Minnesota's Tool for Microscopic Identification (TMI), [42] the Emrath Pigment Database, [43] and the Alaska Fur Database. [44] Though still limited in size, the total number of images is 5410, each paired with a metadata record stored in a JSON file whose string identifiers can be easily converted into a Python dictionary for use in the training framework.

Metadata entries were rewritten using the available source information and hand-edited to use standardized terminology. For example, the string identifier for sample illumination indicates whether the sample was illuminated with transmitted light using an analyzing polarization filter, in which case it is termed "plane-polarized". The term "partially polarized" was used for conditions other than orthogonal (crossed) polarizers, which were labeled as "polarized." For transmitted light, due to inconsistencies in terminology across all of the databases used, we have not differentiated between plane-polarized and true brightfield conditions (the former has a polarization filter in the optical path, while the latter does not), which introduces some inherent noise to our labels. Other illumination types, such as polarized with quarter-wave compensation, darkfield, reflected light, and differential contrast, were added to enhance image diversity and robustness during training.

The images were also hierarchically categorized into classes, subclasses, and descriptions. For example, a human hair fiber would be classified as a "fiber," with a subclass of "hair" and a description of "human." The description was intended to be an unstructured entry, whereas illumination, class, and subclass were structured as drop-down lists of consistent, finite terms. The aim and expectations of this structure were that the model's ease in identifying the sample as a fiber, then a hair, and finally a human hair would diminish as the hierarchy narrowed, providing a logical target for the model to optimize. In addition, we also included, when available, chemical composition, as in the particle's chemical stoichiometric formula, and its refractive index as defining characteristics related to optical microscopy. The JSON file also includes identifiers for staining, magnification, and extended descriptions. Overall, the database lists 19 material types. The distribution is dominated by fibers (n=2,361), pigments (n=1,600), and minerals (n=695), followed by food (n=133), miscellaneous dusts (n=117), and wood sections and fragments (n=114). Fibers account for 44% of the dataset because the Cameo-FRIL library was the most thoroughly compiled and machine-readable dataset available. In future work, and as the database grows, we expect the entries to each material category to become more balanced as images are added.

***Image Preprocessing.*** Prior to running the images through the network, all of them went through a deterministic preprocessing pipeline designed to preserve the per-channel intensity ratio characteristic of optical microscopy data. All inputs were converted to 3-channel RGB images. Pixel values were linearly mapped in the [0,1] space, as the relative magnitudes of color information carry diagnostic information across the transmitted, reflected, and polarized-light modalities present in the corpus. During student training only, the full-resolution image was subjected to independent horizontal and vertical flips (each with probability 0.5) and a uniform in-plane rotation of $\pm 5^{\circ}$ with bilinear resampling, expanded canvas, and black fill when necessary.

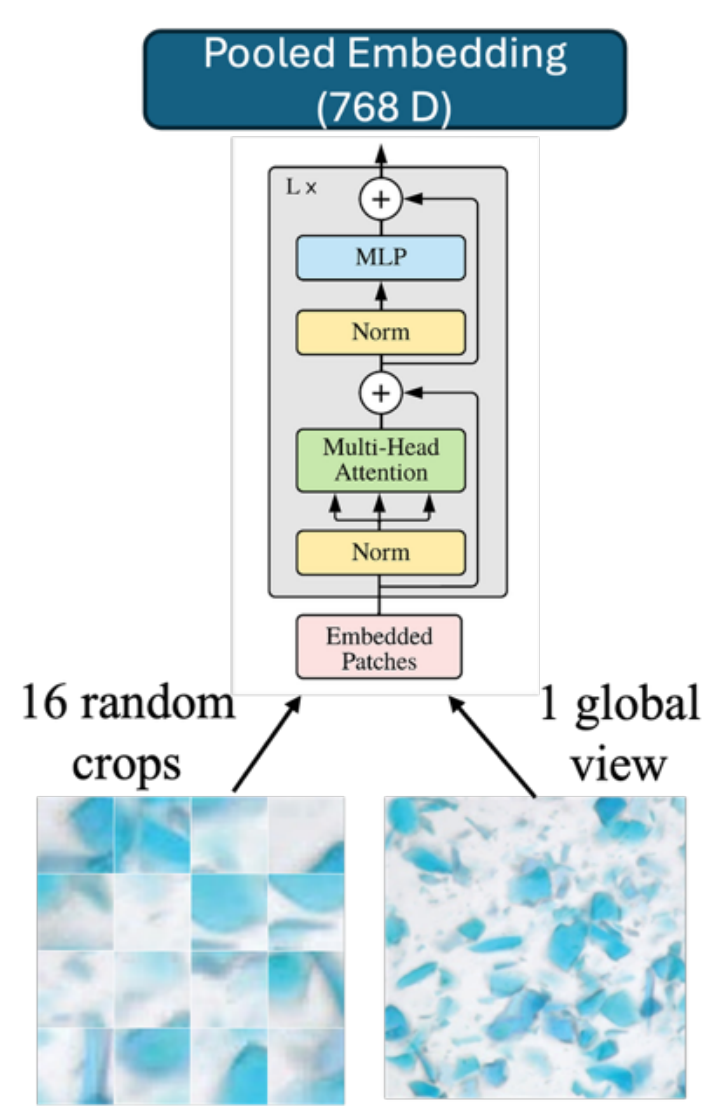

***Figure 3:*** 16 Crops at native pixel resolution and one total view, downsampled and cropped to 224 x 224, are combined to produce "multiview" embeddings.

To match the native CLIP (ViT-B/16) input resolution, each particle image was bilinearly resized to preserve the original aspect ratio, with the short dimension fixed at 224 pixels, then center-cropped to 224 x 224 pixels. As illustrated in Figure 3, this resized image provided a global view of the specimen. In addition to the global view, 16 random crops were extracted at the image's native pixel resolution to retain fine-grained surface detail. The side length $t$ of these crops was adapted to image size on a per-image basis where the smallest images were limited to crops of $16 \times 16$ pixels, which is the ViT-B/16 native tokenizer size. Intermediate sizes are scaled in proportion so that each random crop samples a roughly constant fraction of the specimen's field of view up to a maximum of 224 x 224 pixels. The 1 global view and the 16 random local crops were then fed independently into the image encoder, and the resulting 17 L2-normalized 768-D embeddings were mean-pooled into a single per-image representation for further treatment in downstream stages. These views of the sample at essentially different scales force the model to engage in multi-scale feature fusion. The procedure integrates global context with localized high-resolution information, reducing variance and emphasizing features that are consistent across the field of view. The role of mean pooling across different length scales is to stabilize the embeddings against illumination variability and the inconsistent magnifications used across the dataset.

The image embedding is attached to the text embeddings as follows:

$$E_{fused} = E_{image} \parallel E_{illumination} \parallel \frac{1}{2}(E_{class} + E_{subclass}) \parallel \frac{1}{3}(E_{description} + E_{Composition} + E_{Refractive\ Index}) \quad (3)$$

Where ‖ denotes the end-to-end concatenation of a 768-D image embedding and each 512-length semantic vectors. The combined embedding, $E_{fused}$, with a total length of 2304, serves as the input to our dimensionality-reduction steps using UMAP. The last concatenation, with averaged *Description*, *Composition*, and *Refractive index* fields, depended on whether this information was present (when we had all three metadata, it was an average of three; when we had two of these metadata, it was an average of two; there is never a case when all three are missing). This approach emphasized consistent material properties by using semantic weightings rather than relying solely on illumination or image contrast and averaged out label inconsistencies using multiple descriptors. As a result, samples with similar material characteristics tended to group together in UMAP space.

***Training and Curriculum Labeling Strategy.*** As described above, the student head and ViT backbone were optimized with $l_1$ and CE losses (Eq. 1). To mitigate class imbalance, we used class-balanced (weighted) sampling over the initial 115 pseudo-label classes. The validation cross-entropy remained unweighted, so the reported metrics reflected the natural class distribution.

After an initial warm-up phase to fit the student head, we trained the ViT backbone using progressive unfreezing: (1) only the student MLP head was optimized while the ViT backbone remained frozen; (2) the final layer normalization and projection ("readout") were then unfrozen while all transformer blocks remained frozen; and (3) transformer blocks were subsequently unfrozen in stages until full-model optimization was reached. Learning rates were held constant within each stage, and gradient clipping plus weight decay were used for regularization.

This procedure was then repeated in a second pass, in which the 115 pseudo-labels were combined with a more conventional set of logit targets derived from a tree-structured one-hot encoding of *Illumination*, *Class*, *Subclass*, and *Description*. Labels were restricted to leaf nodes with at least 6 samples, resulting in 266 classes. In a third and final pass, the threshold was relaxed to include only leaf nodes with at least 4 samples, yielding 513 classes. Singleton leaves were excluded from training. This curriculum-labeling strategy progressively sharpens class structure during training, conceptually similar to DINO,[15] in which discriminative pressure is increased over the course of optimization to transition from broad semantic grouping to fine-grained class separation.

***Manuscript writing.*** AI was used in the writing of this manuscript as a copy editor and spellchecker via the Grammarly app (Superhuman Platform Inc., San Francisco). No generative AI content was used in the production of this manuscript.

## 3.0 Results/Discussion

***3.1 Teacher Fusion Produces a Structured Multimodal Manifold.*** In Figure 4, image embeddings from the teacher model's frozen ViT backbone are reduced to 2 dimensions and visualized using UMAP (UMAP-1 on the x-axis and UMAP-2 on the y-axis). The plot features thumbnails of all dataset samples and highlights eleven distinct clusters (marked with Roman numerals), indicating that pre-trained features already encode meaningful contrast and morphological details. For example, Group I includes thin sections of wood. The color of this group is mostly due to staining to reveal microstructures. Groups II-IV contain images of fibers and woven structures, distinguished by their contrast under illumination: lighter backgrounds indicate plane-polarized transmitted light, while darker backgrounds represent differential interference contrast (Group III) or polarized/darkfield illumination (Group IV). Group V (magenta images) shows specimens imaged with a first-order retardation plate in the optical path. Groups VI-XI consist of particle-like materials such as pigments, minerals, dust, and particulates, captured under various polarization conditions. The main factors separating these groups are the illumination modality, which strongly influences background contrast and color, and the morphology, particularly the elongated, high-frequency textures of fibers and textiles. However, the overlapping memberships across several clusters suggest that these pretrained embeddings still lack the domain-specific information needed for reliable particle classification.

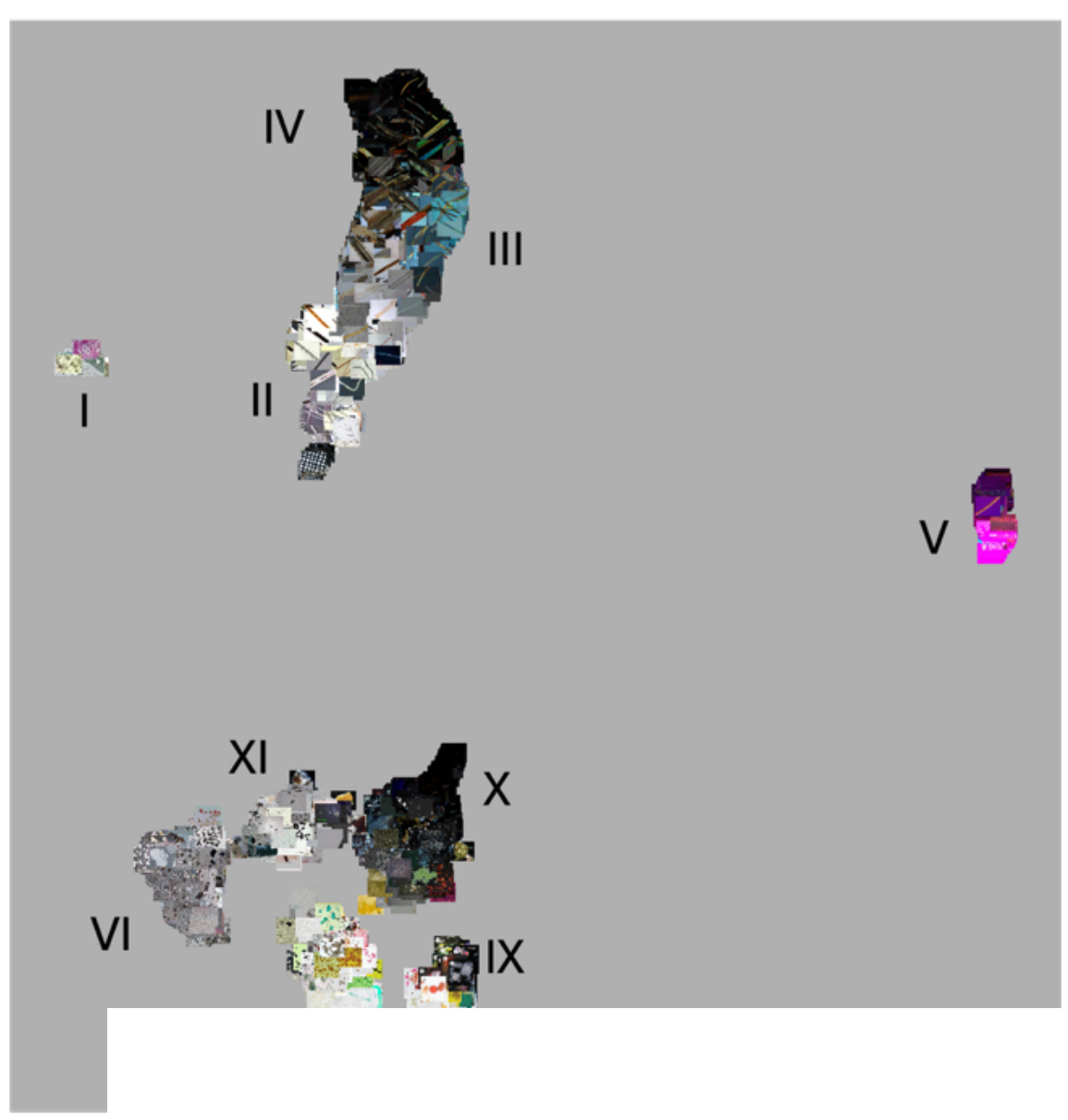

***Figure 4:*** UMAP projection of all dataset embeddings from the pretrained ViT backbone before fine-tuning, shown as dataset thumbnails (UMAP-1, UMAP-2).

To enhance clustering, we combine the image embeddings with each sample's metadata using the concatenation method described in Eq. 3. After testing various fusion strategies, the optimal weighted combination was determined through trial and error, aiming primarily to minimize illumination-based separation in the raw image embeddings while accentuating semantic distinctions for clear groupings. As shown in Figure 5's UMAP visualization, the fused embeddings resolve the broad variability in lighting and morphology into tight, consistent clusters, demonstrating the effectiveness of our approach. Since these clusters are derived from multimodal embedding geometry rather than manually defined taxonomies, they produce a regularization signal that is dense, label-efficient, and closely aligned with perceptual similarities among samples—for example, the consistent grouping of green and blue pigments.

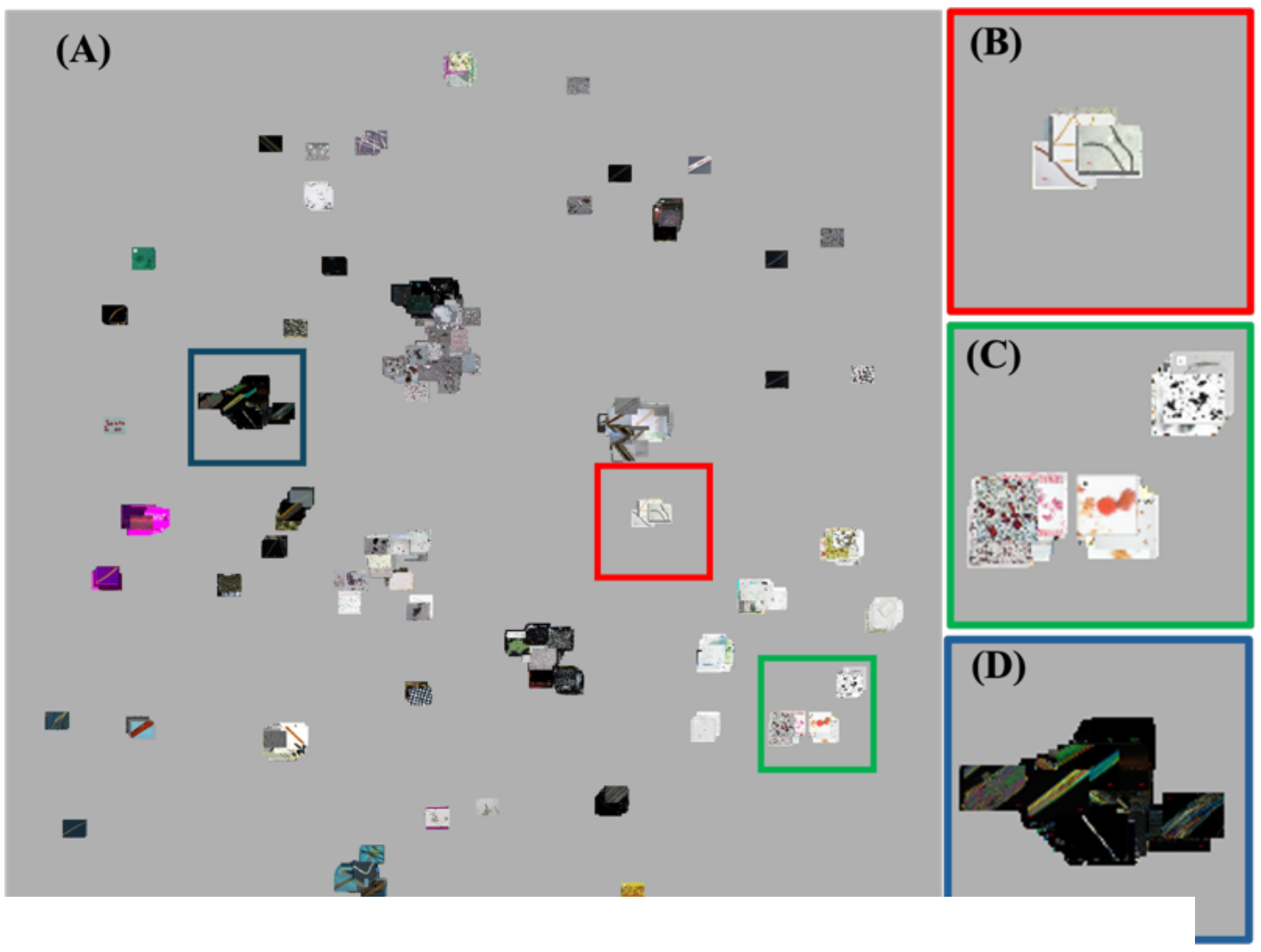

***Figure 5:*** UMAP visualization, (A), shows the teacher's fused embedding space, highlighting specific material groupings: (B) highlights a cluster of synthetic fibers (e.g., nylon, acetate, rayon) under plane-polarized illumination; (C) shows three neighboring clusters corresponding to red, brown, and black pigment particles under plane-polarized illumination. Figure (D) shows a large cluster of natural fibers under crossed-polarized illumination.

We validated these clusters with HDBSCAN (see the supplementary materials), which confirmed that our fusion method produced many well-separated, dense groups of similar materials and identified poorly labeled samples as noise, excluding them from further analysis. Figures 5B-D illustrate this specificity at higher magnification: 5B shows synthetic fibers (nylon, acetate, rayon) under plane-polarized light; 5C displays three neighboring clusters of red, brown, and black pigment particles under plane-polarized illumination; and 5D depicts a large cluster of natural fibers under crossed polarization. These results suggest that off-the-shelf foundation models trained on ImageNet already capture detailed microscopic structures and are broadly adaptable to optical microscopy tasks.

resentations is challenging because effective querying requires prior knowledge of both the image and its metadata, which is never completely available in practice. Therefore, we employ a student-teacher distillation framework, aiming for the student model to internalize the teacher's rich multimodal semantic structure (Figures 5A-D) but focus on a vision-only retrieval task without the need for language inputs. This process emphasizes contrast-specific image features, which are essential for accurate classification and retrieval. In plane-polarized and cross-polarized microscopy, contrast cues often appear as subtle, spatially dispersed patterns arising from light propagation through the material's crystal structure. Training a model to emulate the teacher's detailed features while maintaining alignment with pixel-level ViT features encourages it to remain sensitive to these subtle molecular/crystallographic details.

***3.2 Progressive Unfreezing Transfers the Teacher Manifold to an Image-Only Student.*** To extract diagnostic microstructural features from microscopic images, our training approach preserves much of the generic structure from the pretrained vision backbone. Domain- and task-specific details in our microscopy dataset are mainly learned through a compact, trainable MLP head. This method of reusing foundation-model features and then adapting with lightweight, task-specific heads is now a widely used and effective domain-adaptation strategy, especially for small datasets like the one used in this study. [45-47]

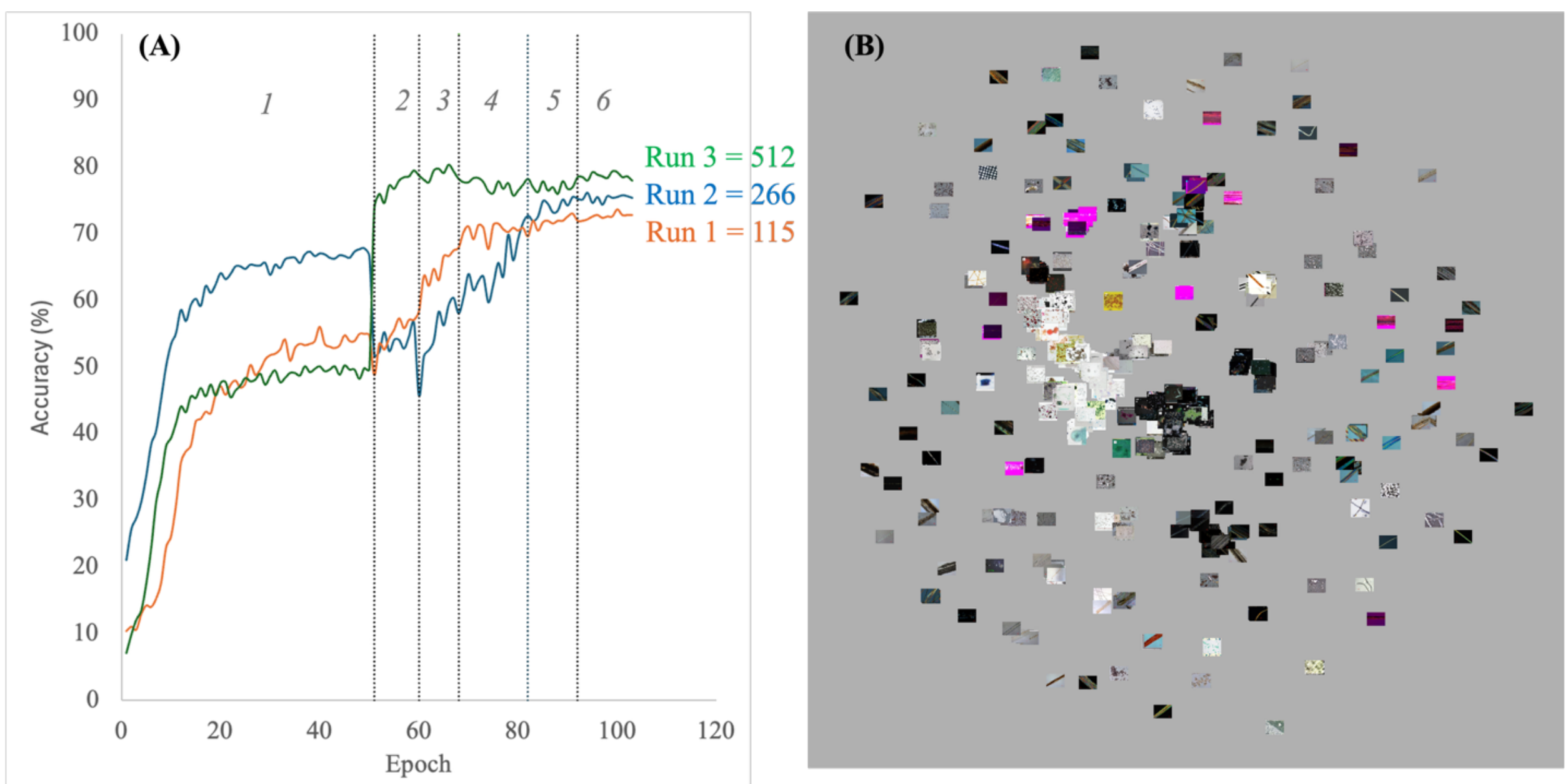

***Figure 6:*** Training the student network uses loss weights with $\beta = 0.11$. (A) First-pass fine-tuning, initialized from pretrained weights and using 115 pseudo-labeled from HDBSCAN, reaches 68% validation accuracy (Run 1). After curriculum expansion to 266 (Run 2) and 513 (Run 3) material classes, reaches 80% validation accuracy. (B) UMAP visualization of the cluster structure learned by the student model after the third pass.

To create a reliable student model, we used a multistage training approach. We first observed that the pretrained ViT already captures useful low- and mid-level visual features—such as texture, edges, and morphology—learned from large natural images, which remain relevant for our domain. Therefore, the MLP head should, in theory, align student features with the fused teacher space without the backbone needing to relearn basic filters. However, the results only partially support this. As shown in Figure 6a, in the first segment, only the head is trained, with validation accuracy initially improving and then stabilizing around 50% during a warm-up phase to seed stable weights (labeled "*1*"). While this is computationally efficient (done on a CPU), it is insufficient for robust retrieval, especially with 115 pseudo-classes.

We then implemented progressive unfreezing and fine-tuning of the ViT to better connect the student head with the backbone features. This process led to about 68% validation accuracy in the initial experiment (run 1 in Figure 6A). The unfreezing process starts with readout-only tuning (second segment), then unfreezes the top two transformer blocks (third segment), followed by staged unfreezing of blocks 8-11, then 4-11, and finally all twelve blocks (0-11), achieving full model optimization at the end.

Discriminative pressure was gradually increased using the curriculum-labeling method outlined in the Methods. The number of material classes grew from 115 (initial pseudo-labels) to 266, then to 513 over three runs (Figure 6A, "Runs 1-3"). This curriculum improved validation accuracy to around 80%, which we consider the baseline and evidence of the training strategy's overall health. After fine-tuning, the student ViT was qualitatively evaluated by analyzing its embedding distribution in UMAP space (Figure 6B). Similarly to the teacher, the student formed multiple dense clusters of visually and semantically similar materials, showing successful transfer of the teacher's manifold structure. The observed difference may be attributed to small errors in training and differences between UMAP runs.

***3.3 Student Embeddings Support Fine-Grained Retrieval.*** The model's ability to retrieve is assessed using two complementary evaluation modes with Facebook AI Similarity Search (FAISS) [48], which finds the nearest examples from a reference embedding index and assigns labels based on those neighbors, thereby measuring in-domain similarity. In our implementation, nearest-neighbor search uses $l_2$ normalized vectors with inner-product ranking and is evaluated using the classic leave-one-out 1-nearest-neighbor (1-NN) metric, where each query is matched to its single closest sample.

Overall Recall@1 is therefore defined as the proportion of queries whose Top-1 retrieved neighbor has the correct label:

$$Recall@1_{overall} = \frac{1}{N}\sum_{i=1}^{N} Recall@1\,\{\hat{y}_i = y_i\} \qquad (4)$$

where $N$ is the number of samples, $y_i$ is the true label of sample $i$, and $\hat{y}_i$ is the predicted label. The indicator term $1\{\hat{y}_i = y_i\}$ equals 1 when the predicted label matches the actual label, else 0.

The second approach, Macro Recall@1 is computed as the unweighted mean of per-group recall:

$$Recall@1_{macro} = \frac{1}{C}\sum_{c=1}^{C} Recall@1_c \qquad (5)$$

where $C$ is the number of unique nodes for the evaluated text field in the dataset tree structure, as detailed in the Methods. Macro Recall@1 is stricter than Overall Recall@1 because rare and frequent classes are weighted equally, so this metric is not overly influenced by classes with large membership. Results for both metrics are reported in Table 1.

| Field Description | Tree Nodes | $Recall@1_{overall}$ (%) | $Recall@1_{macro}$ (%) |
|---|---|---|---|
| Illumination | 11 | 98.8 | 93.4 |
| Class | 21 | 98.6 | 82.7 |
| Subclass | 91 | 94.0 | 63.1 |
| Description | 824 | 74.9 | 28.8 |

**Table 1:** Recall Accuracy of the Student Model

The results in Table 1 present recall for *Illumination*, *Class*, *Subclass,* and *Description* on the labeled portions of the dataset. Both Overall Recall@1 and Macro Recall@1 are high across all tree nodes except for *Description*. This is expected because labels like *polarized* for the illumination modality, *fiber* for the class, and *synthetic fiber* for the subclass contain cues that a human microscopist would naturally consider as prior contextual knowledge about the specimen type and imaging conditions. Nevertheless, these metrics strongly indicate that the student model is identifying these microscopically relevant features from the pixel data.

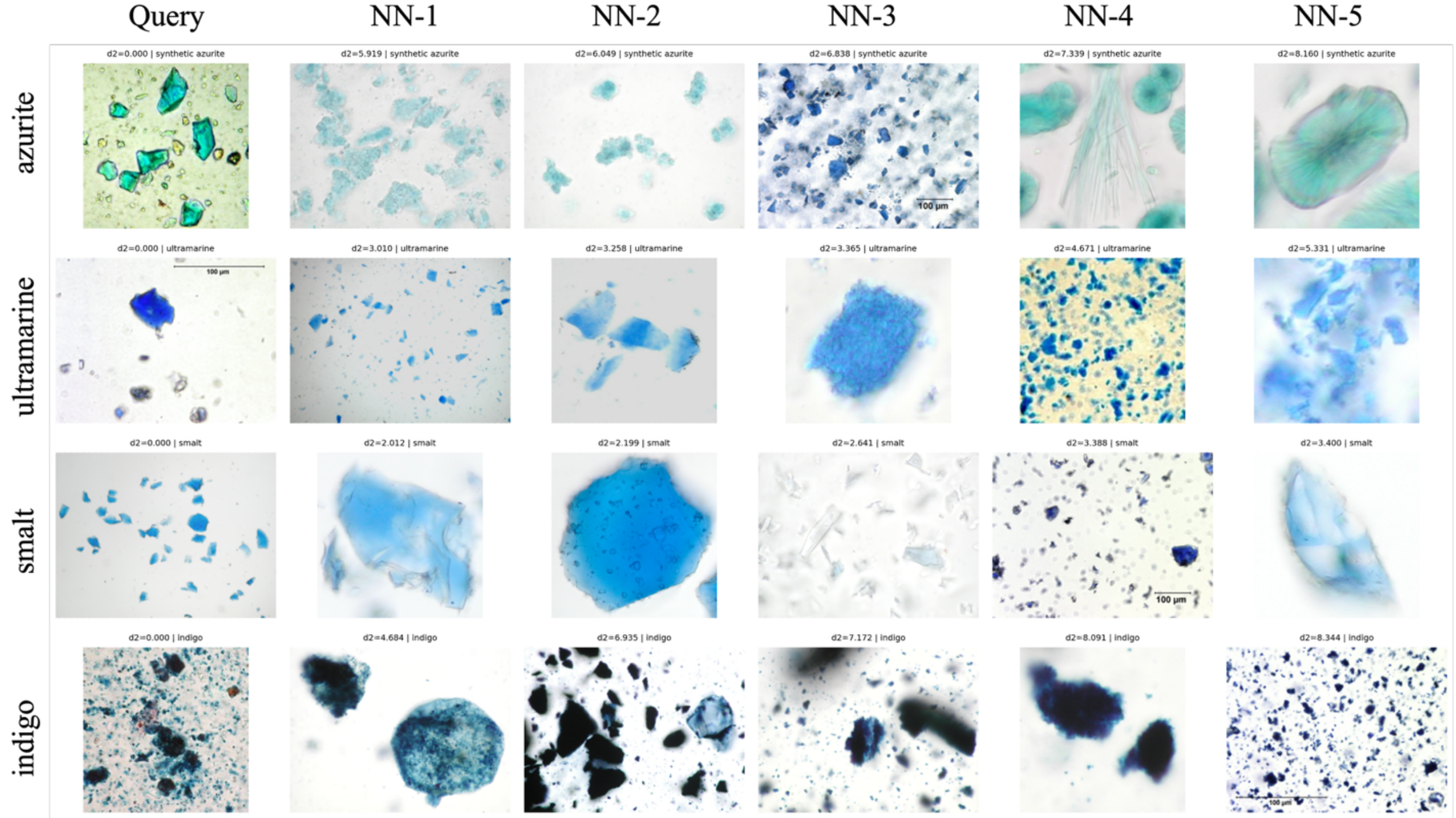

***Figure 7:*** Various blue pigments under plane polarized light. Note that the particles are not at uniform magnification but still group according to their descriptive identity.

The *Description* field labels are more diagnostic of fine-grained performance: Overall Recall@1 is nearly 75%, indicating strong retrieval for common, well-supported groups, while Macro Recall@1 is only ~29%, revealing weaker performance on rare or small groups. This gap highlights a clear dataset imbalance that could be reduced by increasing the number of sample images in small groups, which will be the focus of future work. Even so, ~29% Top-1 remains far above chance, at about 238 times the odds of a random guess under the current label cardinality. At Top-10 recall, *Description* performance improves (Macro Recall@10 ~38%), while a very high Macro Recall@10 (~96.7%) is achieved when evaluated with 513 classes used for training. It is important to note that the 824 *Description* leaves are derived from the overall dataset (excluding singletons), so all groups with support ≥ 2 are present. The 513 leaves have a different granularity threshold with a support ≥ of 4, which explains the observed differences in the two results.

To illustrate the specificity of recall performance, we include two examples: one with blue pigments (Figure 7) and another with typical synthetic fibers (Figure 8). The rows of the figures represent the *Description*-level nodes of our dataset organization, while the columns show, respectively, the query image and its top five nearest neighbors (NN-1 through NN-5). In all these cases, the query and its top five closest neighbors share the same *Description* labels, confirming the robustness of the recall.

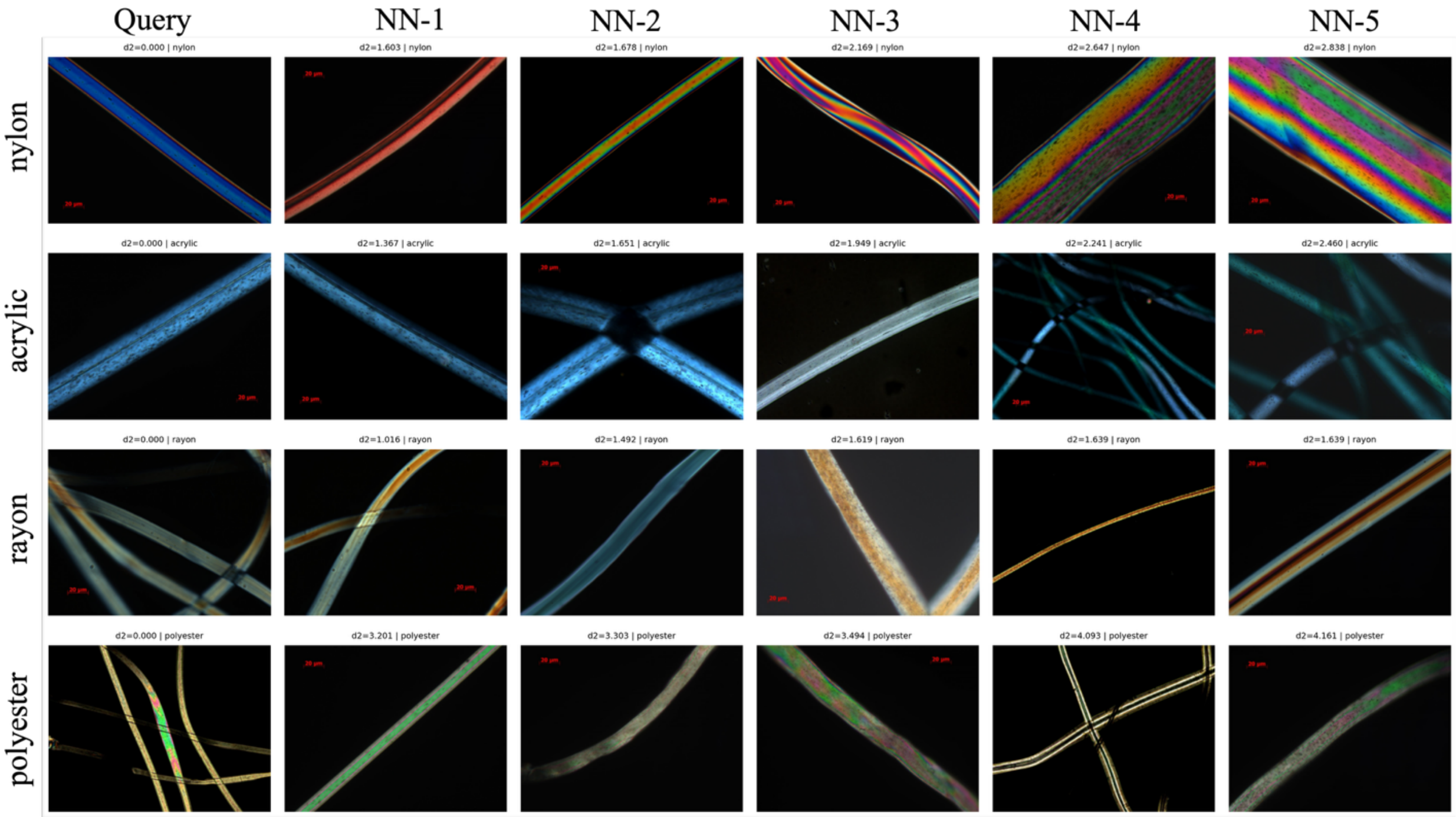

***Figure 8:*** Various undyed and transparent synthetic fibers under crossed polarized light. The fibers are not shown at a uniform scale. The differences in thickness can account for the interreference color dissimilarities, especially in the nylon in the first row.

Both of these examples typically involve complex identification challenges. Distinguishing between a variety of blue pigments or a selection of transparent polymer threads has traditionally required sophisticated analytical tools such as Raman or FTIR spectrometers. Interestingly, we can attain this level of specificity using only legacy microscopic data. Key features of these queries include their robustness to magnification changes. For example, in Figure 7, the nearest neighbors of the pigment images are not limited to the same magnification. Zoomed-in or -out views still match the query, thanks to the multiview pipeline (outlined in the Methods), which was designed to reduce this bias. Similarly, plane-polarized illumination reveals differences in background contrast and color jitter. An example of this jitter is the yellow tint in the azurite query sample, yet the recall remains stable despite this noise. The polymer fiber images in Figure 8, taken under crossed-polarized light, display interference colors related to their refractive indices and thicknesses, as explained by light-matter interaction physics. Even fibers of different sizes—visible via the scale bars, especially among nylon and polyester—cluster together despite differences in interference colors. For instance, the thin blue nylon query sample appears perceptually distinct yet still clusters with other nylon examples. These differences suggest that the model has also learned interference-color cues related to polarized light. Understanding why these features activate and how the model maintains such robustness is the focus of the next section on interpretation.

***3.4 Illumination Semantics Are Encoded as Spatial Latent Contrast.*** The student-teacher architecture naturally facilitates the investigation of how semantic structures are encoded within the vision model. The teacher's embedding consists of a sequence of semantically meaningful segments: an image segment followed by text-derived segments for illumination, class/subclass, and description. We hypothesize that, through distillation, this block-level semantic segmentation is transferred to the student's latent space and related back to the image's pixel values during training through backpropagation. To verify this hypothesis in a straightforward and interpretable manner, we focus on the illumination segment, since each sample is categorized into one of a finite set of illumination modalities in the database.

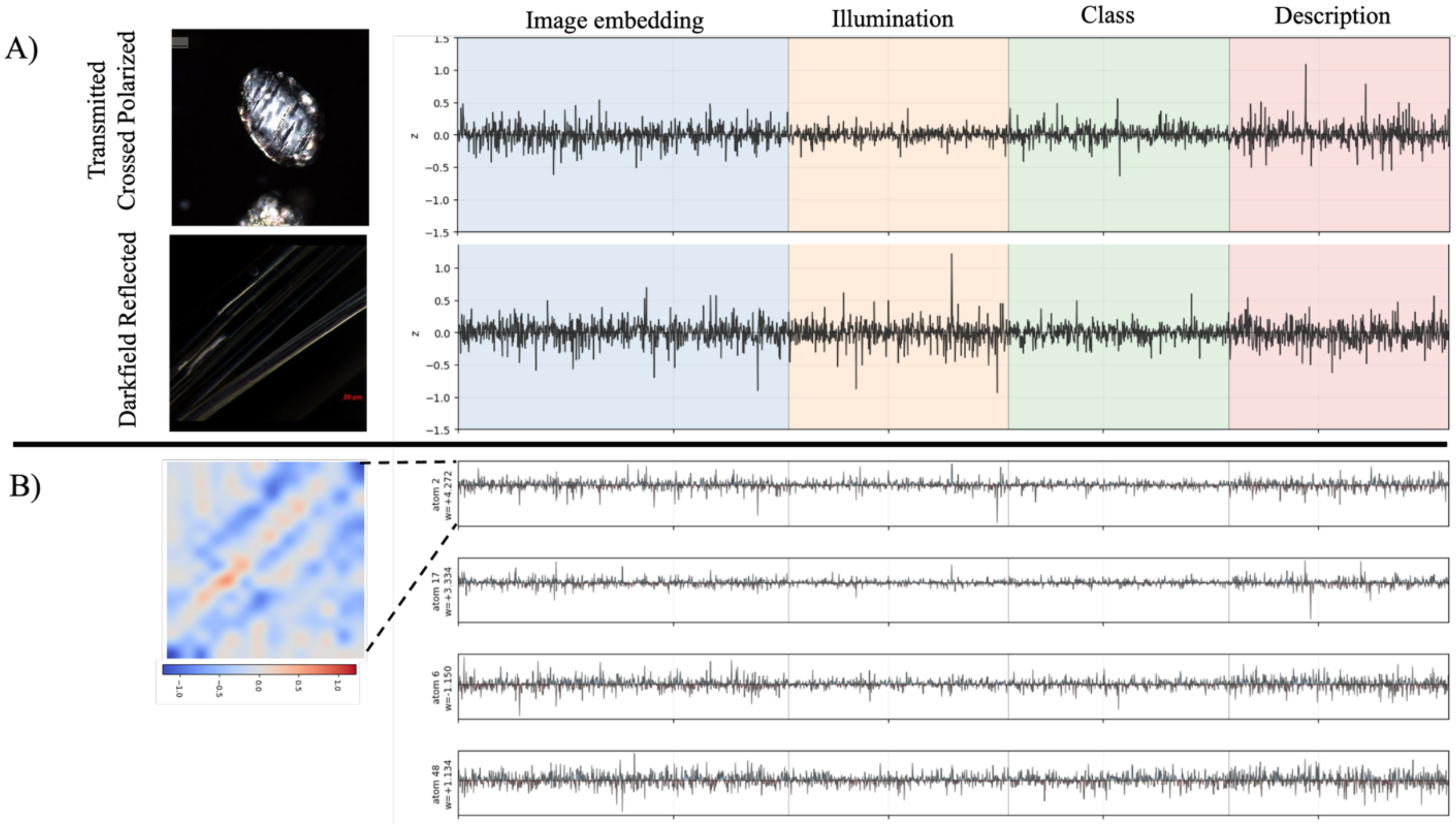

***Figure 9:*** Block-level divisions of the student embeddings show different energies. (A) shows two illumination modalities with dark-colored backgrounds for a feldspar mineral and a plant fiber, with transmitted crossed-polarized light and darkfield reflected, respectively. The energy in the illumination block for both is quite different: darkfield has high energy and polarized has low energy. (B) shows top atoms from dictionary learning and the spatial strength of one of the atoms with respect to the darkfield fiber image.

For each student embedding $\boldsymbol{z} \in \mathbb{R}^{2304}$, we compute the fraction of total embedding energy contained in the illumination block:

$$f_{illum}(z) = \frac{\|z_{768:1279}\|_2}{\|z_{0:2303}\|_2} = \frac{\sum_{j=768}^{1279} z_j^2}{\sum_{j=0}^{2303} z_j^2} \quad (6)$$

Here $z_j$ is the value of the $j-th$ dimension of the student embedding. The numerator measures the $l_2$ norm of the illumination block. The denominator measures the $l_2$ norm of the full 2304-dimensional embedding. Ranking samples by $f_{illum}$ identifies cases in which the student allocates relatively high or low energy to illumination-related dimensions.

Using this metric, we observe that the sample with the highest illumination-block energy fraction is a plant fiber imaged under darkfield reflected illumination, while the sample with the lowest fraction is a mineral imaged under transmitted crossed polarization (see Figure 9A). This pattern is consistent across the dataset: high-energy fractions are associated with darkfield-reflected images, whereas low fractions correspond to transmitted crossed-polarized images. Transmitted plane-polarized and other illumination types fall between these two, each with distinct impulse patterns for their respective illumination concepts. Since a pixel mechanism based solely on contrast is unlikely to reliably distinguish between the two dark backgrounds—where pixel intensities are near zero—this suggests that the student has learned to differentiate darkfield reflected from transmitted crossed-polarized illumination by leveraging the semantic structure provided by the teacher during distillation.

This raises an important question about how the student captures contrast-related features from a teacher model that encoded this information only at the semantic level. Likewise, if the near-zero background pixels alone do not contain enough information to define the illumination modality, where is the concept of darkfield or transmitted crossed polarization spatially encoded in the image? To address this, we decomposed the 2304-dimensional student embeddings $\boldsymbol{z}$ using a sparse dictionary model:

$$\boldsymbol{z} \approx \mathbf{D}\boldsymbol{\alpha} = \sum_{k=1}^{K} \alpha_k \, \mathbf{d}_k \qquad (7)$$

where $\mathbf{D}$ is a dictionary, whose columns $\mathbf{d}_k$ are the base representations of the semantic features inherent to our dataset, and $\boldsymbol{\alpha}$ (with elements $\alpha_k$) is the sparse coefficient vector and $K$ is the number of columns in the dictionary. Each $\mathbf{d}_k$ is the k-th dictionary atom, a learned base representation of a visual feature, and $\alpha_k$ is its scalar coefficient, indicating how strongly that feature is present in the embedding. Sparsity means that for each embedding, only a few coefficients $\alpha_k$ are non-zero. Each atom represents a recurring direction in the student's latent space, and each coefficient measures how strongly that direction contributes to reconstructing a given embedding. Because the coefficients are sparse, each image embedding is reconstructed by only a small subset of atoms.

We used K-SVD[49] to learn a 200-atom dictionary ($K$ = 200 in eq. 7) and lasso-based sparse coding[50] to decompose all 2304-dimensional student embeddings. For the darkfield reflected image shown in Figure 9B, the largest positive reconstruction weights corresponded to atoms 2, 17, 6, and 48. Across multiple darkfield reflected samples, Atom 2 repeatedly appears with a large positive coefficient, suggesting that this atom captures a latent direction strongly associated with the darkfield reflected condition.

To locate where this latent direction appears spatially, we extracted the final ViT patch tokens, forming a 14 × 14 grid over the image. Each token was processed through the student head to generate a local 2304-dimensional embedding. Atom 2 was then independently fitted to each local embedding, creating a spatial coefficient map in Figure 9b with both positive and negative values. Positive coefficients indicate local alignment with the same atom direction as in the global embedding, while negative coefficients indicate the opposite. Although the global embedding assigns Atom 2 a strong positive coefficient, the spatial map shows a more detailed relationship, with positive coefficients clustered in the dark background and negative ones on the fibers. This suggests that the darkfield reflected condition is not solely encoded by the specimen or background but by a contrastive latent pattern involving both a positive background component and an opposing specimen-related component. This indicates that illumination information, provided only as semantic supervision to the teacher, was transferred into the student as a spatial contrast pattern. Instead of merely reproducing a global label, the student learned a representation where both background and specimen contrast features jointly encode the darkfield-reflected condition. While this is just one of many possible examples, these data demonstrate that semantic steering of a foundation model is possible, thereby enhancing the model's expressiveness and its ability to distinguish microscopic specimens under different lighting conditions without access to the clear class labels required by classical supervised training.

## Conclusions

This work shows that weak multimodal metadata can be distilled into a practical model that uses only images to identify the contents of microscopic particles and fibers. By transferring knowledge from a combined image-text teacher model to a student Vision Transformer (ViT), the system learned embeddings that maintain semantic relationships while relying only on images at inference. The resulting features achieved high nearest-neighbor recall for detailed descriptive labels, suggesting that the student captured microscopy-relevant features rather than dataset-specific labels. An interpretability analysis indicates that semantic concepts, such as illumination modality, are not just stored as global labels but can also appear as spatially organized contrast patterns within the student embedding. These results demonstrate that teacher-student semantic distillation is a useful method for turning weakly labeled microscopy archives into searchable, interpretable representations for analyzing heterogeneous particles and fibers.

## Author Contributions

M.W. and A.K. conceived of the project, secured funding, and supervised the overall direction of the work. M.W. wrote most of the manuscript based on the experimental work of S.S., A.O., and M.B.M., all of whom also contributed to the writing. L.L. and K.N. helped supervise the students. N.E. contributed to the writing of the manuscript, supplied images for training of the ViT, and provided consultation on PLM.

## Acknowledgements

This work was initially funded by a grant from the National Endowment for the Humanities prior to 2024, when the project's funding was halted by the Trump Administration and is to be reinstated.

## Competing Interests

All authors declare no financial or non-financial competing interests.

## Data Availability

The datasets produced and analyzed in this study are not publicly accessible due to proprietary reasons. However, they can be obtained from the corresponding author upon reasonable request.

## Code Availability

The underlying code for this study is not publicly available but may be made available to qualified researchers upon reasonable request from the corresponding author.